\documentclass[letterpaper]{article} 
\usepackage{d4ir}  
\usepackage{times}  
\usepackage{helvet}  
\usepackage{courier}  
\usepackage[hyphens]{url}  
\usepackage{graphicx} 
\usepackage{natbib}  
\usepackage{caption} 
\usepackage{algorithm}
\usepackage{algorithmic}

\usepackage{amsmath}
\usepackage{textcomp}
\usepackage{xcolor}

\usepackage{array}
\usepackage{textcomp}
\usepackage{url}
\usepackage{verbatim}
\usepackage{booktabs}
\usepackage{multirow}
\usepackage{color} 

\usepackage{float}
\usepackage{subfig}
\usepackage{subfloat}
\usepackage{overpic}
\usepackage{arydshln}

\usepackage{newfloat}
\usepackage{listings}
\DeclareCaptionStyle{ruled}{labelfont=normalfont,labelsep=colon,strut=off} 
\floatstyle{ruled}
\newfloat{listing}{tb}{lst}{}
\floatname{listing}{Listing}
\title{ Data-free Distillation with Degradation-prompt Diffusion for Multi-weather Image Restoration }
\author{
    Pei Wang\textsuperscript{\rm 1}\equalcontrib, 
    Xiaotong Luo\textsuperscript{\rm 1}\equalcontrib,
    Yuan Xie\textsuperscript{\rm 2},
    Yanyun Qu\textsuperscript{\rm 1}\thanks{Corresponding author.}
}
\affiliations{
    \textsuperscript{\rm 1}School of Informatics, Xiamen University, Xiamen, China\\
    \textsuperscript{\rm 2} School of Computer Science and Technology, East China Normal University, Shanghai, China \\

    pwang0219@stu.xmu.edu.cn, xiaotongtt123@gmail.com, yxie@cs.ecnu.edu.cn, yyqu@xmu.edu.cn
}

\usepackage{bibentry}
\nocopyright

\begin{document}

\maketitle

\begin{abstract}
Multi-weather image restoration has witnessed incredible
progress, while the increasing model capacity and expensive data acquisition impair its applications in memory-limited devices.
Data-free distillation provides an alternative for allowing to learn a lightweight student model from a pre-trained teacher model without relying on the original training data. 
The existing data-free learning methods mainly optimize the models with the pseudo data generated by GANs or the real data collected from the Internet.
However, they inevitably suffer from the problems of unstable training or domain shifts with the original data. 
In this paper, we propose a novel Data-free Distillation with Degradation-prompt Diffusion framework for multi-weather Image Restoration (D4IR). 
It replaces GANs with pre-trained diffusion models to avoid model collapse and incorporates a degradation-aware prompt adapter to facilitate content-driven conditional diffusion for generating domain-related images. 
Specifically, a contrast-based degradation prompt adapter is firstly designed to capture degradation-aware prompts from web-collected degraded images. 
Then, the collected unpaired clean images are perturbed to latent features of stable diffusion, and conditioned with the degradation-aware prompts to synthesize new domain-related degraded images for knowledge distillation. 
Experiments illustrate that our proposal achieves comparable performance to the model distilled with original training data, and is even superior to other mainstream unsupervised methods. 
\end{abstract}

%
%
%
%

\section{Introduction}

Multi-weather image restoration (MWIR) aims to recover a high-quality image from a degraded input (e.g., haze, rain), which can be used in autonomous driving, security monitoring, etc. 
Nowadays, MWIR \cite{li2022all,cui2024adair} has made significant progress relying on 
the rapid development of computing hardware and the availability of massive data. 
In actual scenarios, the increasing model complexity may impair its application on resource-constrained mobile vehicular devices. 
As a widely used technique, Knowledge Distillation (KD) \cite{luo2021boosting,DBLP:conf/aaai/ZhangSZSZ24} is often adopted for model compression.
However, the original training data is unavailable for some reasons, e.g., transmission constraints or 
privacy protection. 
Meanwhile, due to the variability of weather conditions, access to large-scale and high-quality datasets containing all weather conditions can be both difficult and expensive.
Therefore, it is necessary to develop data-free learning methods to compress existing IR models for adapting to different edge devices and more robust to various adverse weather conditions. 

\begin{figure}[t]
    \centering
    \includegraphics[width=\linewidth]{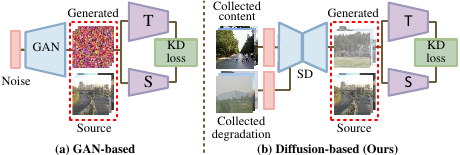}
    \caption{The schematic diagram comparison of the data-free distillation methods for MWIR. (a) GAN-based methods 
    : directly map the pure noise to the original data domain, while (b) our diffusion-based method 
 synthesizes images with separate content and degradation information.} 
    \label{intro_fig}
    \vspace{-10pt}
\end{figure}

Data-free knowledge distillation \cite{lopes2017data} paves such a way to obtain lightweight models without relying on the original training data. 
Its core concern is how to acquire data similar to the training data. 
The existing methods mainly achieve knowledge transfer by generating pseudo-data based on generative adversarial networks (GANs) \cite{chen2019data, zhang2021data} or collecting trust-worth data from the Internet \cite{chen2021learning, tang2023distribution}. 
However, these methods mainly focus on high-level tasks, lacking sufficient exploration in low-level image restoration for pixel-wise dense prediction. 

Recently, a few studies \cite{zhang2021data, wang2024data} have explored data-free learning for image restoration. 
However, there are still two underlying limitations. 
Firstly, they all adopt the GAN-based framework, which often faces unstable training and complex regularization hyperparameter tuning. 
Secondly, they use pure noise as input to generate pseudo-data that generally lack clear semantic and texture information. It is crucial for low-level vision tasks.
Although collecting data from the Internet can avoid the problem, it would inevitably face domain shift from the original data, which is difficult to solve for MWIR unlike simple perturbations based on class data statistics \cite{tang2023distribution} in image classification. 


In order to mitigate the above issues, we advocate replacing GANs with a pre-trained conditional diffusion model and equipping it with degradation-aware prompts to generate domain-related images from content-related features. 
On the one hand, the diffusion models can avoid mode collapse or training instability of GANs and are superior in covering the modes of distribution \cite{nichol2021improved}. 
On the other hand, by training on large-scale datasets, many conditional diffusion models (e.g., Stable Diffusion (SD) \cite{Rombach2021HighResolutionIS} ) demonstrate exceptional ability in creating images that closely resemble the content described in the prompts. 
Especially, some methods \cite{dong2023prompt, liu2024diff} resort to the powerful prior of these pre-trained models and introduce trainable adapters to align the internal learned knowledge with external control signals for task-specific image generation. 

In this paper, we propose a novel \textbf{D}ata-free \textbf{D}istillation with \textbf{D}egradation-prompt \textbf{D}iffusion for multi-weather \textbf{I}mage \textbf{R}estoration (D4IR). 
As shown in Fig. \ref{intro_fig}, unlike previous GAN-based data-free learning methods \cite{wang2024data} for MWIR, 
our D4IR separately extracts degradation-aware and content-related feature representations from the unpaired web-collected images with conditional diffusion to better approach the source distribution. 
It aims to shrink the domain shift between the web-collected data and the original training data.

Specifically, our D4IR includes three main components: degradation-aware prompt adapter (DPA), content-driven conditional diffusion (CCD), and pixel-wise knowledge distillation (PKD). 
DPA and CCD are jointly utilized to generate degraded images close to the source data.
For DPA, a lightweight adapter is employed to extract degradation-aware prompts from web-collected low-quality images, which employs contrastive learning to effectively learn diverse degradation representations across different images. 
For CCD, the encoded features of web-collected clean images are perturbed to latent samples by forward diffusion, and then conditioned with the degradation-aware prompts for synthesizing data near the source distribution under the degradation reversal of the teacher model. 
With the newly generated images, the student network could be optimized to mimic the output of the teacher network through PKD. 
Experiments illustrate that our proposal achieves comparable performance to distill with the original training data, and is even superior to other mainstream unsupervised methods. 

In summary, the main contributions are four-fold: 
\begin{itemize}
    \item We propose a novel data-free distillation method for MWIR, which aims to break the restrictions on expensive model complexity and data availability.
    \item We design a contrast-based adapter to encode degradation-aware prompts from various degraded images, and then embed them into stable diffusion.
    \item We utilize the diffusion model to capture the latent content-aware representation from clean images, which combines the degradation-aware prompts to generate data that is more consistent with the source domain.
    \item Extensive experiments demonstrate that our method can achieve comparable performance to the results distilled with the original data and other unsupervised methods. 
\end{itemize}

\section{Related Works}
\subsection{Multi-weather Image Restoration} 
MWIR can be divided into single-task specific models for deraining \cite{DBLP:conf/aaai/0004CLL24, DBLP:journals/tip/WangJWRZL24}, dehazing \cite{DBLP:conf/aaai/0018PLD0W24}, desnowing \cite{DBLP:conf/mm/ZhangJWZNZ23, quan2023image}, and multi-task all-in-one IR models \cite{li2022all, cui2024adair}. 
Based on the physical and mathematical models, many MWIR methods \cite{DBLP:journals/tmm/LiLZKY23} attempt to decouple degradation and content information from the training data. 
For example, DA-CLIP~\cite{luocontrolling} adapts the controller and fixed CLIP image encoder to predict high-quality feature embeddings for content and degradation information. 
Recently, transformer-based models \cite{song2023vision} have been introduced into low-level tasks to model long-range dependencies, significantly improving performance. 
Restormer \cite{zamir2022restormer} designs a efficient multi-head attention and feed-forward network to capture global pixel interactions. 
Though these methods have made powerful performance, the substantial storage space and computational resources make them challenging to deploy on resource-constrained edge devices.

Moreover, due to the difficulty in obtaining large-scale paired degraded-clean images, many methods use unpaired data to achieve unsupervised IR based on techniques like GANs~\cite{wei2021deraincyclegan}, contrastive learning~\cite{ye2022unsupervised, wang2024ucl}, etc. 
Unlike these methods, our proposal combines disentanglement learning and stable diffusion to generate data closer to the source domain for KD. 

\begin{figure*}[t]
    \centering
    \includegraphics[width=0.9\textwidth]{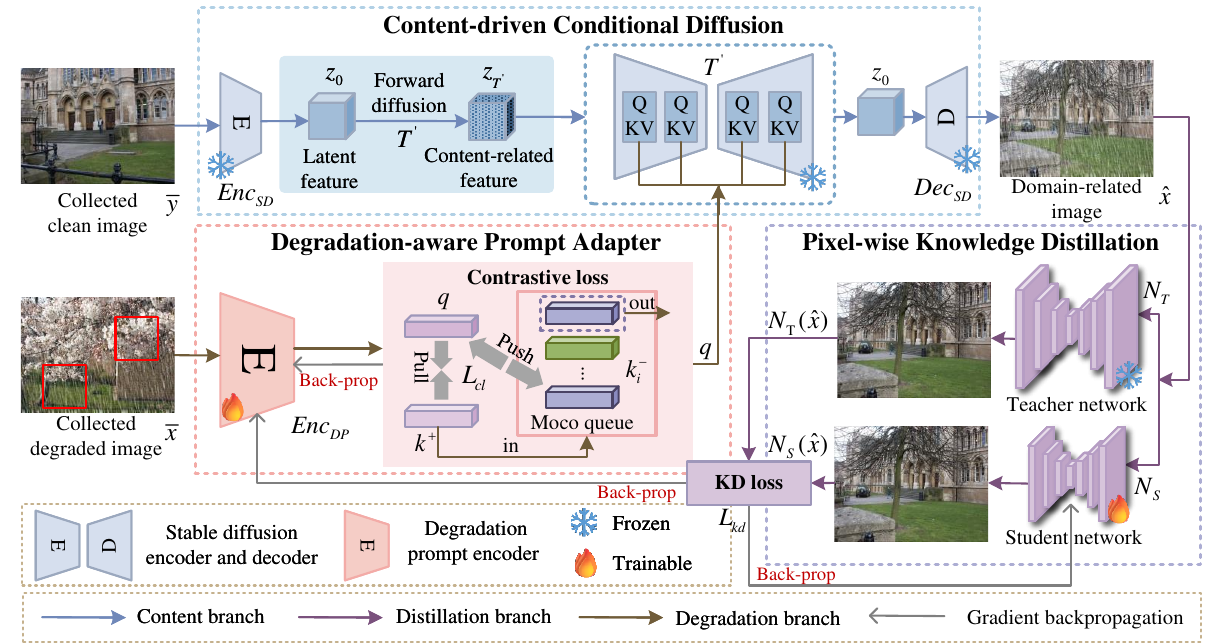}
    \vspace{-5pt}
    \caption{
    The overall framework of our proposed D4IR. It separately extracts degradation-aware and content-related features from unpaired web-collected images to guide SD in synthesizing the source domain related images for knowledge distillation. 
    }
    \label{framework}
    \vspace{-10pt}
\end{figure*}

\subsection{Data-free Knowledge Distillation \label{2.2}}
Existing data-free distillation methods can be roughly classified into three types.
Firstly, the methods \cite{lopes2017data, nayak2019zero} reconstruct training samples in the distillation process with the ``metadata" preserved during training. However, they are less feasible when only the pre-trained teacher model is accessible due to the necessity of ``metadata". 
Secondly, the methods \cite{micaelli2019zero, fang2019data} optimize GANs to generate data similar to the distribution of original training data by a series of task-specific losses. 
DAFL \cite{chen2019data} distills the student network by customizing one-hot loss, information entropy loss, and activation loss based on classification features. 
DFSR \cite{zhang2021data} introduces data-free distillation to image SR and designs the reconstruction loss with bicubic downsampling to achieve performance comparable to the student network trained with the original data. 
DFMC \cite{wang2024data} adopts a contrastive regularization constraint to further improve model representation based on DFSR for MWIR. 
The last methods \cite{chen2021learning, tang2023distribution} optimize with web-collected data and try to address the distribution shift between collected data and original training data. KD3 \cite{tang2023distribution} selects trustworthy instances based on classification predictions and learning the distribution-invariant representation. 

\subsection{Conditional Diffusion Models}
To achieve flexible and controllable generation, conditional diffusion methods combine the auxiliary information (e.g., text \cite{saharia2022photorealistic}, image \cite{zhao2024uni}, etc.) to generate specific images. 
In particular, Stable Diffusion (SD) \cite{Rombach2021HighResolutionIS} successfully integrates the text CLIP \cite{radford2021learning} into latent diffusion. 

Given the efficiency of foundation models such as SD, most recent methods \cite{dong2023prompt, liu2024diff} resort to their powerful prior and introduce trainable prompts to encode different types of conditions as guidance information. 
For example, T2I-Adapter \cite{mou2023t2i} enables rich controllability in the color and structure of the generated results by training lightweight adapters to align the internal knowledge with external control signals according to different conditions. 
Diff-Plugin \cite{liu2024diff} designs a lightweight task plugin with dual branches for a variety of low-level tasks, guiding the diffusion process for preserving image content while providing task-specific priors. 

\section{Proposed Method}

\subsection{Preliminary}
\noindent\textbf{Notation and Formulation.} 
Formally, given the pre-trained teacher network $N_T(\cdot)$, knowledge distillation (KD) aims to learn a lightweight student network $N_S(\cdot)$ by minimizing the model discrepancy $dis(N_T, N_S)$. With the original training data $D = \{(x_i, y_i)\}_{i=1}^{|D|}$ (``$|\cdot|$" is the data cardinality, $x_i$ and $y_i$ are the degraded image and clean image), traditional KD is usually achieved by minimizing the following loss: 
\begin{equation}
    L_{kd}(N_S) = \frac{1}{|D|} \sum_{i=1}^{|D|}[{\parallel N_T(x_i) - N_S(x_i)\parallel}_2]
\end{equation} 

\noindent\textbf{Problem Definition.} 
In practice, the original training data $D$ may be inaccessible due to transmission or privacy limitations, which hinders efficient model training. 
That means only the pre-trained teacher model is available.
Therefore, our D4IR aims to address two significant issues for data-free KD:
(1) how to capture the data for model optimization; (2) how to achieve effective knowledge transfer.

Technically, data-free KD methods simulate $D$ with generated pseudo-data or web-collected data. 
To efficiently synthesize the domain-related images to the original degraded data for MWIR, we first analyze the mathematical and physical models \cite{su2022survey} used in traditional IR method. 
The general formulation of the degraded image $Y$ is assumed to be obtained by convolving a clean image $X$ with a fuzzy kernel $B$ and further adding noise $n$ as follows:
\begin{equation}
    Y = X * B + n \label{eq1}
\end{equation}
where $*$ denotes convolution operation. Inspired by disentangled learning \cite{DBLP:journals/tmm/LiLZKY23}, we consider decoupling the low-quality images as degradation-aware ($B$, $n$) and content-related information ($X$) from web-collected degraded images $\bar{D}_X = \{\bar{x}_i \}_{i=1}^{|\bar{D}_X|}$ and unpaired clean images $\bar{D}_Y =\{ \bar{y}_i\}_{i=1}^{|\bar{D}_Y|}$ to facilitate the pre-trained SD model to generate source domain-related degraded images.  

\subsection{Method Overview}
As illustrated in Fig. \ref{framework}, our method consists of three main components: degradation-aware prompt adapter (DPA), content-driven conditional diffusion (CCD), and pixel-wise knowledge distillation (PKD). These parts are collaboratively worked to generate data close to the source domain so as to achieve data-free distillation of MWIR.
 
First, DPA includes a lightweight learnable encoder $Enc_{DP}$, which is used to extract degradation-aware prompts $Enc_{DP}(\bar{x})$ from the collected degraded images $\bar{x}$. 
To learn task-specific and image-specific degradation representations across various images, $Enc_{DP}$ is trained with contrastive learning \cite{he2020momentum}, i.e., the features of patches from the same image ($q$, $k^+$) are pulled closer to each other and pushed away from ones of other images ($k_i^-$). 


Then, CCD performs the diffusion process from the perturbed latent features $z_{T^{'}}$ of the collected clean images $\bar{y}$, which is designed to relieve the style shift between the original data and the images generated by frozen stable diffusion \cite{Rombach2021HighResolutionIS} starting from random noise.
Moreover, $z_{T^{'}}$ is conditioned with the degradation-aware prompts $Enc_{DP}(\bar{x})$ for synthesizing new domain-related images $\hat{x}$. 

Finally, PKD is conducted with the generated images $\hat{x}$. Without loss of generality, the student network is optimized with a pixel-wise loss $L_{kd}$ between its output $N_S(\hat{x})$ and the one of teacher network $N_T(\hat{x})$. 
Note that $L_{kd}$ is utilized to simultaneously optimize $N_S(\cdot)$ and $Enc_{DP}$. 
It aims to filter the degradation types domain-related to the original data from large-scale collected images for contributing to KD. 

\subsection{Degradation-aware Prompt Adapter \label{sec3.3}} 

As previously discussed, the degradation-aware prompt adapter (DPA) aims to extract the degradation representations that help the student network learn from the teacher network with web-collected low-quality images. 
To achieve this, the adapter needs to satisfy the following conditions. 

First, DPA expects to effectively learn diverse degradation representations across different images while focusing on the task-specific and image-specific degradation information that distinguishes it from other images for the input image. 
Therefore, we adopt contrastive learning \cite{Hnaff2019DataEfficientIR, Chen2020ASF} to optimize DPA to pull in the same degradation features and push away irrelevant features. 

Specifically, we randomly crop two patches $\bar{x}_q$ and $\bar{x}_{k^+}$ from the collected degraded image $\bar{x}$, which are considered to contain the same degradation information. 
Then, they are passed to a lightweight encoder $Enc_{DP}$ with three residual blocks and a multi-layer perceptron layer to obtain the corresponding features $q=Enc_{DP}(\bar{x}_q)$ and $k^+=Enc_{DP}(\bar{x}_{k^+})$. We treat $q$ and $k^+$ as query and positive samples. 
On the contrary, the features $k_i^-=Enc_{DP}(\bar{x}_{k_i^-})$ of the patches $\bar{x}_{k_i^-}$ cropped from other images are viewed as negative samples. All negative sample features are stored in a dynamically updated queue of feature vectors from adjacent training batches following MoCo \cite{he2020momentum}. Thus, the contrastive loss $L_{cl}$ can be expressed as:
\begin{equation}
    L_{cl}(Enc_{DP}) = -\log {\frac{exp( q \cdot k^+ / \tau)}{\sum_{i=1}^K exp( q \cdot k_i^- / \tau)}}
    \label{eqcl}
\end{equation} 
where $\tau$ is a temperature hyper-parameter set as $0.07$ \cite{he2020momentum} and $K$ denotes the number of negative samples. 

Second, DPA needs to extract domain-related prompts to guide the diffusion model in synthesizing images that facilitate knowledge transfer. If we only use Eq. (\ref{eqcl}) to optimize $Enc_{DP}$, the resulting prompts may overlook the degradation differences between the web-collected data and the original training data. 
This implies that DPA might only capture degradation features across different input images, leading to a distribution shift from the original data. 
To address this, we employ the distillation loss $L_{kd}$ between the outputs of the student model and teacher model to simultaneously optimize the degradation prompt encoder and the student model. 

Replacing the text prompt encoder in the pre-trained SD model, we employ the DPA to align the internal knowledge prior with external encoded degradation-aware prompts by the cross-attention module~\cite{Rombach2021HighResolutionIS} for generating images toward specific degradation-related images: 
\begin{equation}
    Attention(Q,K,V) = softmax(\frac{QK^T}{\sqrt{d}})\cdot V
\end{equation}
Q, K, and V projections are calculated as follows: 
\begin{equation}
\begin{aligned}
    Q = W_Q^{(i)} \cdot \varphi_i(z_t), \enspace & K = W_K^{(i)} \cdot Enc_{DP}(\bar{x}), \\
    V = W_V^{(i)} &\cdot Enc_{DP}(\bar{x})
\end{aligned}
\end{equation}
where $\varphi_i(z_t)$ denotes the intermediate representation of the UNet in SD. $W_Q^{(i)}$, $W_K^{(i)}$, and $W_V^{(i)}$ are projection matrices frozen in SD. $d$ is the scaling factor \cite{vaswani2017attention}. 
\subsection{Content-driven Conditional Diffusion}

According to the degradation prompts, the diffusion models still cannot generate domain-related images. This is because they inevitably suffer from the content and style differences against the real images without specifying the content of the images. 
Therefore, it is necessary to address the content shift from the original degraded data while preserving the realism of the collected images.

Inspired by SDEdit \cite{meng2021sdedit}, we choose the noised latent features $z_{T^{'}}$ encoded from the collected clean image $\bar{y}$ instead of the random noise to synthesize domain-related images with realism. Specifically, we first encode the web-collected clean images $\bar{y}$ into latent representations $z_0$ by the encoder $Enc_{SD}$ frozen in SD via $z_0 = Enc_{SD}(\bar{y})$.

Then, we replace the initial random Gaussian noise with the ${T^{'}}$-step noised features $z_{T^{'}}$ of the latent features $z_0$ as the input to the diffusion model:  
\begin{equation}
    z_t = \sqrt{\bar{\alpha}_t}z_0 + \sqrt{1-\bar{\alpha}_t}\epsilon_t, \enspace t={T^{'}}
    \label{t}
\end{equation}
where $\bar{\alpha}_t$ is the pre-defined schedule variable \cite{song2020denoising}, $\epsilon_t \sim N(0,1)$ is the random noise, ${T^{'}} = \lambda * T $, $T$ is the total number of sampling steps in the diffusion model, and $\lambda \in [0,1]$ is a hyper-parameter indicating the degree of injected noise. 

With the learned conditional denoising autoencoder $\epsilon_\theta$, the pre-trained SD can gradually denoise $z_{T^{'}}$ to $z_0$ conditioned with the degradation-aware prompts $Enc_{DP}(\bar{x})$ via
\begin{equation}
\begin{aligned}
    z_{t-1} = &\sqrt{\bar{\alpha}_{t-1}}(\frac{z_t-\sqrt{1-\bar{\alpha}_{t}}\epsilon_\theta(z_t, t , Enc_{DP}(\bar{x}))}{\sqrt{\bar{\alpha}_{t}}}) 
    \\
    & + \sqrt{1-\bar{\alpha}_{t}}\cdot\epsilon_\theta(z_t, t , Enc_{DP}(\bar{x}))
\end{aligned}
\end{equation} 

Finally, the decoder $Dec_{SD}$ reconstructs the image $\hat{x}$ from the denoised latent feature $z_0$ as $\hat{x} = Dec_{SD}(z_0)$.

As the noised input $z_{T^{'}}$ to the diffusion model retains certain features of the real image $\bar{y}$, the generated image $\hat{x}$ closely aligns in style with the real image. More importantly, by starting from the partially noised features of the collected clean images, the pre-trained SD model can generate images $\hat{D} = \{(\hat{x}_i)\}_{i=1}^{|\hat{D}|}$ that reflect the content and degradation characteristics of the original training data, when conditioned with degradation-aware prompts $Enc_{DP}(\bar{x})$. 

\subsection{Pixel-wise Knowledge Distillation}
Considering that image restoration focuses on pixel-level detail in an image, we calculate the distillation loss $L_{kd}$ by the pixel-wise distance between the outputs of the student network and the teacher network as:
\begin{equation}
    L_{kd}(N_S, Enc_{DP}) = \frac{1}{|\hat{D}|} \sum_{i=1}^{|\hat{D}|}[{\parallel N_T(\hat{x}_i) - N_S(\hat{x}_i)\parallel}_2]
    \label{kd}
\end{equation}
where $\hat{x_i}$ denotes the synthesized images. For better generalization, we simply provide a simple way to conduct distillation, and other KD losses are also encouraged. 

Note that the distillation loss is used to optimize both the student network and the degradation prompt adapter. Therefore, the whole objective function is formulated as:
\begin{equation}
    L(N_S, Enc_{DP}) = L_{kd}(N_S, Enc_{DP}) + \gamma \cdot L_{cl}(Enc_{DP})
    \label{l}
\end{equation}
where $\gamma$ is a regularization coefficient to balance the distillation loss and the contrastive loss.




\section{Experiments} 

\begin{table*}[t]
\centering
\small
\vspace{-7pt}
\begin{tabular}{clcccc}
    \toprule
    \textbf{Type} & \textbf{Method} & \textbf{Params(M) $\downarrow$} & \textbf{PSNR(dB) $\uparrow$} & \textbf{SSIM $\uparrow$} \\
    \hline
    \multirow{7}{*}{Unsupervised} & CUT~\cite{park2020contrastive} & 14.14 & 23.01 & 0.800 \\
    & DeraincycleGAN~\cite{wei2021deraincyclegan} & 28.86 & 31.49 & 0.936 \\
    & DCD-GAN~\cite{chen2022unpaired} & 11.4 & 24.06 & 0.792 \\
    & NLCL~\cite{ye2022unsupervised} & 0.63 & 27.77 & 0.644 \\
    & Cycle-Attention-Derain \cite{chen2023cycle} & / $^{\mathrm{a}}$ & 29.26 & 0.902 \\
    & Mask-DerainGAN~\cite{wang2024mask} & 8.63 & 31.83 & 0.937 \\
    \hline
    Teacher &AirNet~\cite{li2022all} & 8.52 & 34.90 & 0.966 \\
    Student & Half-AirNet & 4.26 & 30.88 & 0.924 \\
    \hline
    \multirow{4}{*}{KD}& Data (Half-AirNet) & 4.26 & 29.12 & 0.883 \\
    & DFSR \cite{zhang2021data} & 4.26 & 28.39 & 0.859\\
    & DFMC \cite{wang2024data} & 4.26 & 29.59 & 0.882 \\
    & D4IR (Ours) & 4.26  & 30.03 & 0.906 \\
    \bottomrule
    \multicolumn{5}{l}{$^{\mathrm{a}}$ The codes of them are not officially available.}
\end{tabular}
\vspace{-5pt}
\caption{Quantitative results of D4IR and other methods for image deraining on Rain100L.}\label{tab_derain}
\end{table*}

\begin{figure*}[t]
\small
\centering
        \begin{minipage}{0.19\linewidth}
		\centerline{\includegraphics[width=\textwidth]{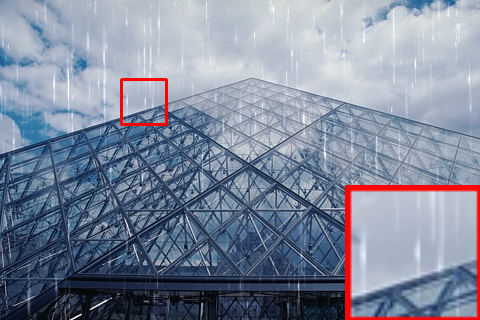}}
		\centerline{Rainy}
        \end{minipage}
        \begin{minipage}{0.19\linewidth}
		\centerline{\includegraphics[width=\textwidth]{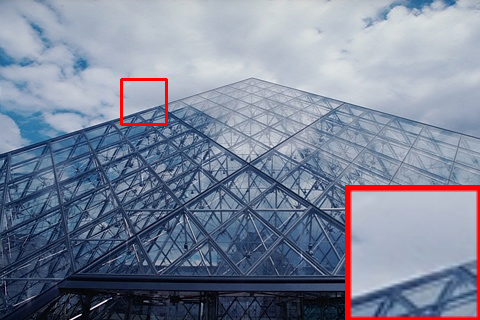}}
		\centerline{DerainCycleGAN}
        \end{minipage}
        \begin{minipage}{0.19\linewidth}
		\centerline{\includegraphics[width=\textwidth]{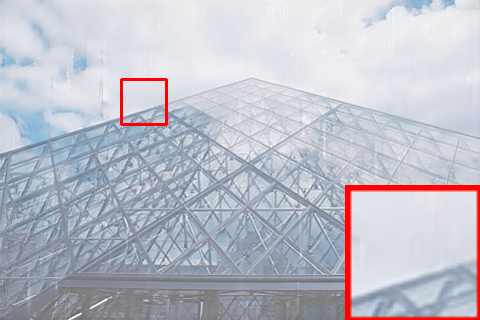}}
		\centerline{NLCL}
        \end{minipage}
        \begin{minipage}{0.19\linewidth}
        \centerline{\includegraphics[width=\textwidth]{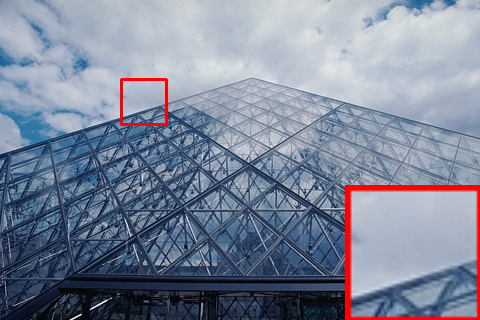}}
		\centerline{Teacher}  
        \end{minipage}
        \begin{minipage}{0.19\linewidth}
		\centerline{\includegraphics[width=\textwidth]{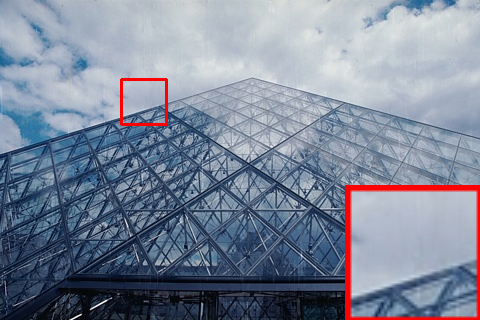}}
		\centerline{Student}  
        \end{minipage}

        \begin{minipage}{0.19\linewidth}
		\centerline{\includegraphics[width=\textwidth]{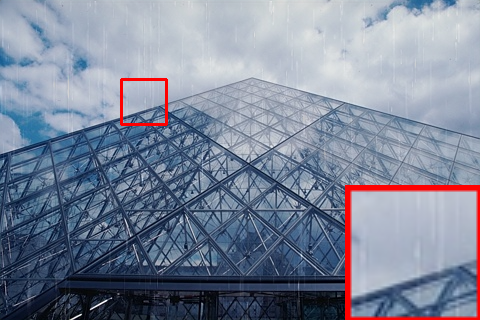}}
		\centerline{Data}
        \end{minipage}
        \begin{minipage}{0.19\linewidth}
		\centerline{\includegraphics[width=\textwidth]{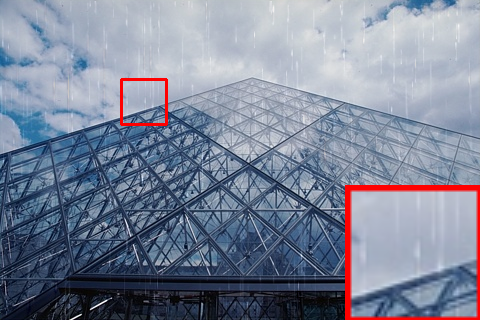}}
		\centerline{DFSR}
        \end{minipage}
        \begin{minipage}{0.19\linewidth}
		\centerline{\includegraphics[width=\textwidth]{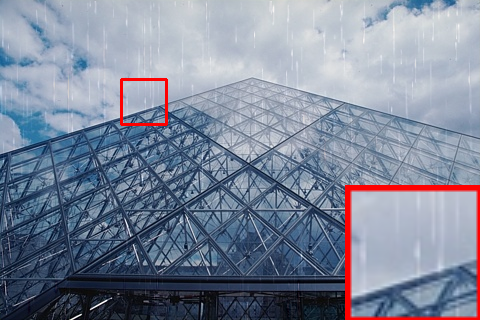}}
		\centerline{DFMC}
        \end{minipage}
       \begin{minipage}{0.19\linewidth}
		\centerline{\includegraphics[width=\textwidth]{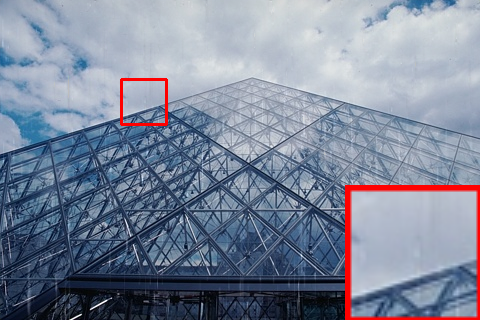}}
		\centerline{Ours}
        \end{minipage}
        \begin{minipage}{0.19\linewidth}
		\centerline{\includegraphics[width=\textwidth]{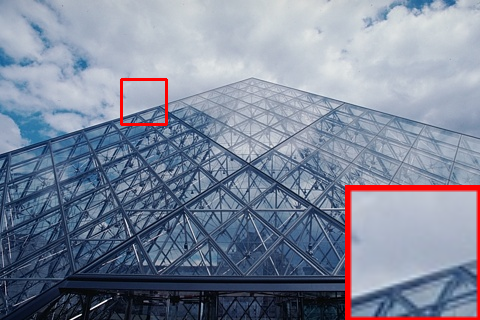}}
		\centerline{GT}
        \end{minipage}  
	\centering
 \vspace{-5pt}
	\caption{Visual comparisons of D4IR and other methods for image deraining on Rain100L. Zoom in for a better view.
 }
	\label{fig_rain}
    \vspace{-10pt}
\end{figure*}


\subsection{Experimental Settings}
\textbf{Datasets}. Following the previous work in high-level tasks \cite{tang2023distribution}, we introduce the web-collected data to synthesize data near the original distribution. Specifically, our datasets are as follows: 

\textit{1) Original Training Datasets:} Here, we mainly consider the common weather following the representative AirNet \cite{li2022all}. The teacher networks are trained on Rain100L \cite{yang2017deep} for deraining, the Outdoor Training Set (OTS) \cite{li2018benchmarking} for dehazing, and Snow100K \cite{liu2018desnownet} for desnowing. 

\textit{2) Web-Collected Datasets:} 
For image draining, we employ the training images from the large-scale deraining dataset Rain1400 \cite{fu2017removing} with $12,600$ rainy-clean image pairs. 
For image dehazing, we adopt the training images from RESIDE \cite{li2018benchmarking} with $72,135$ outdoor and $13,990$ indoor hazy-clean image pairs. 
For image desnowing, we set the training images from the Comprehensive Snow Dataset (CSD) \cite{chen2021all} with $8,000$ snowy-clean image pairs. 
Note that the paired images are randomly shuffled during training to reach an unpaired configuration.

\textit{3) Test Datasets:} 
Following the common test setting for different weather image restoration, we adopt Rain100L \cite{yang2017deep}, Synthetic Objective Testing Set (SOTS) \cite{li2018benchmarking}, and the test datasets of Snow100K for image deraining, dehazing and desnowing, respectively. 

\noindent\textbf{Implementation Details.} 
We employ the pre-trained AirNet as the teacher network and then halve the number of feature channels to obtain the student network. 
The initial learning rates of the student network $N_S(\cdot)$ and the degradation prompt encoder $Enc_{DP}$ are set as $1\times10^{-3}$ and $1\times10^{-5}$, respectively, which are decayed by half every $15$ epoch. 
Adam optimizer is used to train D4IR with $\beta_{1}=0.9$ and $\beta_{2}=0.999$. 
The specific sampling step of the latent diffusion~\cite{Rombach2021HighResolutionIS} is $70$. 
During training, the input RGB images are randomly cropped into $256\times256$ patches and the batch size is set following AirNet. 
To ensure the training stability, we first train $N_S(\cdot)$ and $Enc_{DP}$ together as Eq. (\ref{l}) for $50$ epochs, and then with the distillation loss as Eq. (\ref{kd}) for $150$ epochs. 
Besides, the hyperparameter $\lambda$ in Eq. (\ref{t}) and the trade-off parameter $\gamma$ in Eq. (\ref{l}) are set as $0.5$ and $0.5$, respectively (the analysis is shown in the supplementary material). 
All experiments are conducted in PyTorch on NVIDIA GeForce RTX 3090 GPUs.

\noindent\textbf{Evaluation Metrics.} Peak signal-to-noise ratio (PSNR) \cite{huynh2008scope} and structural similarity (SSIM) \cite{wang2004image} are utilized to evalute the performance of our method. 
Besides, the parameters are used to evaluate model efficiency.

\begin{table*}[t]
\centering
\small
\vspace{-5pt}
\begin{tabular}{clcccccc}
    \toprule
    \textbf{Type} & \textbf{Method} & \textbf{Params(M) $\downarrow$} & \textbf{PSNR(dB) $\uparrow$} & \textbf{SSIM $\uparrow$} \\
    \hline
    \multirow{7}{*}{Unsupervised} & YOLY~\cite{li2021you}  & 32.00 & 19.41 & 0.833 \\
    & RefineDNet~\cite{zhao2021refinednet} & 65.80	& 24.23	& 0.943 \\
    & D4~\cite{yang2022self} & 10.70 & 25.83 & 0.956 \\
    & VQD-Dehaze ~\cite{yang2023visual} & 0.23 & 22.53 & 0.875 \\
    & IC-Dehazing~\cite{gui2023illumination} & 15.77 & 24.56 & 0.929 \\
    & UCL-Dehaze~\cite{wang2024ucl} & 22.79 & 25.21 & 0.927 \\
    & ADC-Net~\cite{wei2024robust} & 26.56 & 25.52 & 0.935 \\
    \hline
    Teacher & AirNet~\cite{li2022all} & 8.93 & 25.75 & 0.946 \\
    Student & Half-AirNet & 4.46 & 25.69 & 0.944 \\
    \hline
    \multirow{4}{*}{KD}& Data (Half-AirNet) & 4.46 & 25.63 & 0.945 \\
    & DFSR~\cite{zhang2021data} & 4.46 & 21.33 & 0.890 \\
    & DFMC~\cite{wang2024data} & 4.46 & 21.96 & 0.900 \\
    & D4IR (Ours) & 4.46 & 25.67 & 0.946 \\ 
    \bottomrule
\end{tabular}
\vspace{-5pt}
\caption{Quantitative results of D4IR and other methods for image dehazing on SOTS.}\label{tab_dehaze}
\end{table*}

\begin{figure*}[t]
\small
\centering
        \begin{minipage}{0.16\linewidth}
		\vspace{1pt}
		\centerline{\includegraphics[width=\textwidth]{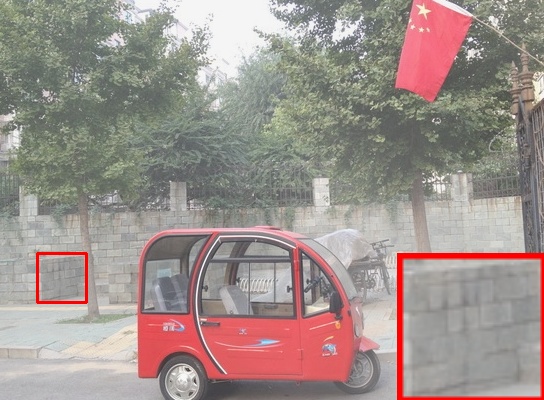}}
		\centerline{Hazy}
        \end{minipage}
        \begin{minipage}{0.16\linewidth}
		\centerline{\includegraphics[width=\textwidth]{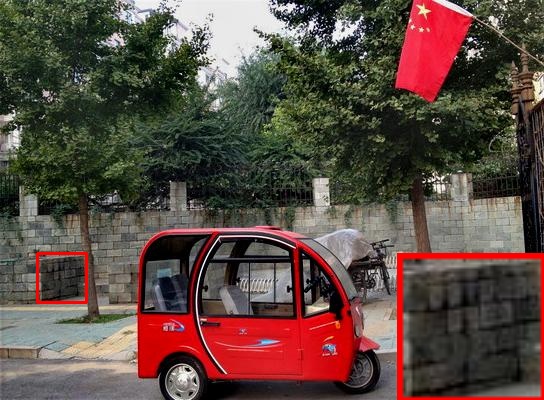}}
		\centerline{YOLY}
        \end{minipage}
        \begin{minipage}{0.16\linewidth}
		\centerline{\includegraphics[width=\textwidth]{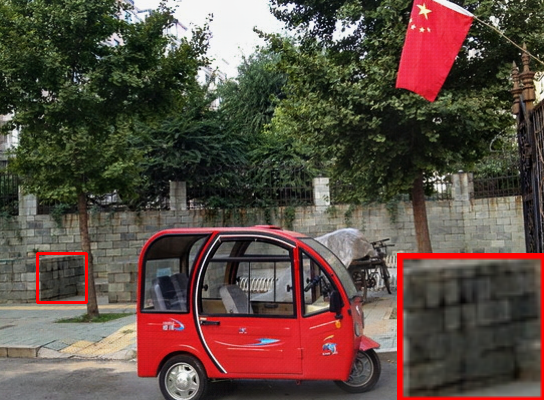}}
		\centerline{RefineDNet}
        \end{minipage}
        \begin{minipage}{0.16\linewidth}
		\centerline{\includegraphics[width=\textwidth]{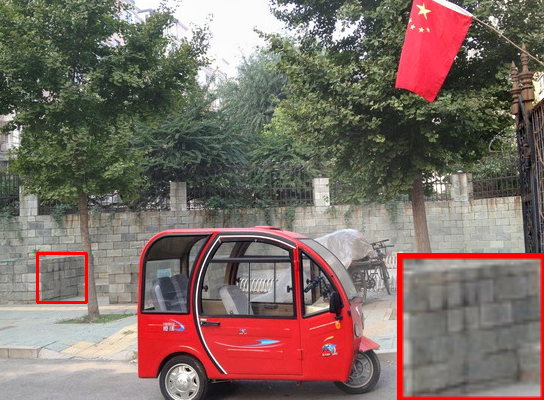}}
		\centerline{D4}
        \end{minipage}
        \begin{minipage}{0.16\linewidth}
        \centerline{\includegraphics[width=\textwidth]{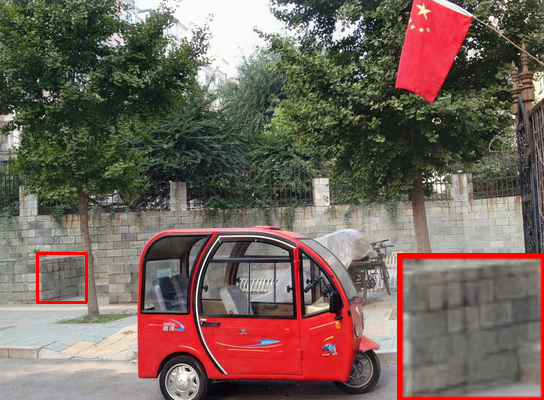}}
		\centerline{Teacher}  
        \end{minipage}
        \begin{minipage}{0.16\linewidth}
		\centerline{\includegraphics[width=\textwidth]{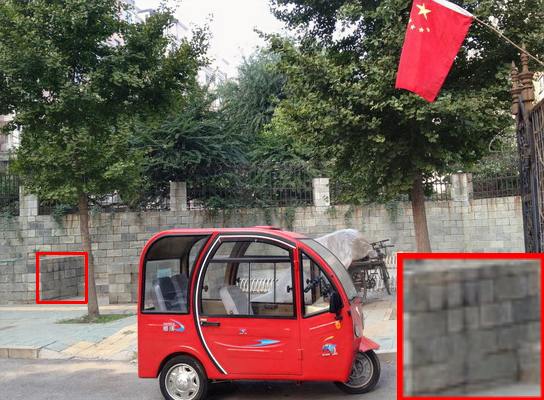}}
		\centerline{Student}  
        \end{minipage}

        \begin{minipage}{0.16\linewidth}
		\vspace{1pt}
		\centerline{\includegraphics[width=\textwidth]{result_fig/dehaze_paste/hazy.jpg}}
		\centerline{Hazy}
        \end{minipage}
        \begin{minipage}{0.16\linewidth}
		\centerline{\includegraphics[width=\textwidth]{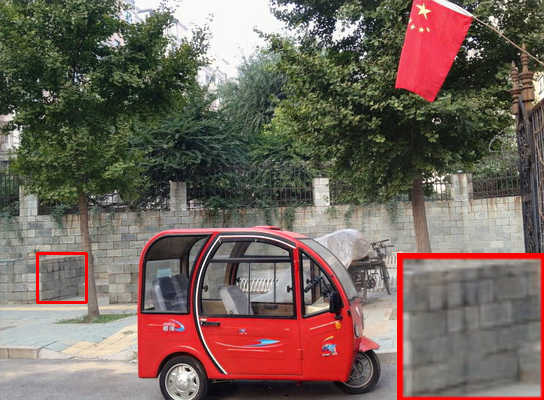}}
		\centerline{Data}
        \end{minipage}
        \begin{minipage}{0.16\linewidth}
		\centerline{\includegraphics[width=\textwidth]{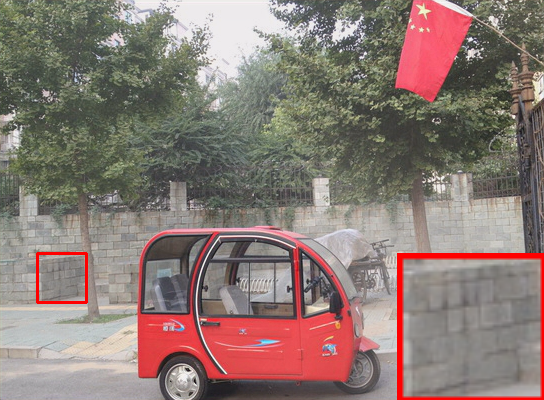}}
		\centerline{DFSR}
        \end{minipage}
        \begin{minipage}{0.16\linewidth}
		\centerline{\includegraphics[width=\textwidth]{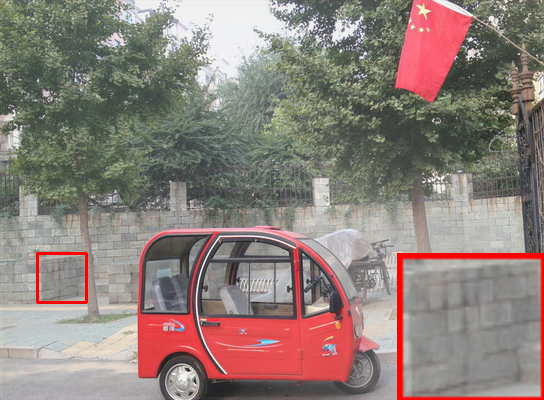}}
		\centerline{DFMC}
        \end{minipage}
       \begin{minipage}{0.16\linewidth}
		\centerline{\includegraphics[width=\textwidth]{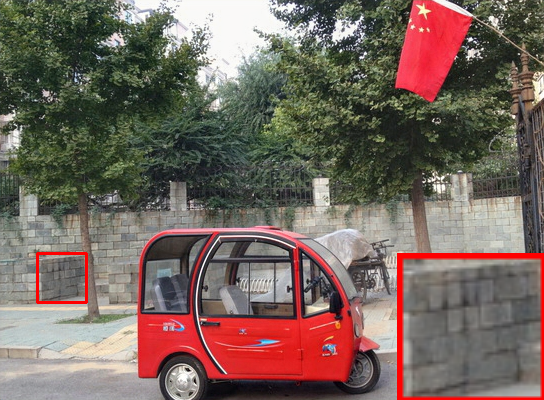}}
		\centerline{Ours}
        \end{minipage}
        \begin{minipage}{0.16\linewidth}
		\centerline{\includegraphics[width=\textwidth]{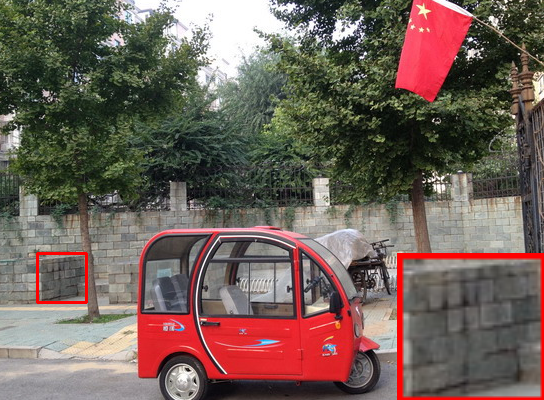}}
		\centerline{GT}
        \end{minipage}  
        
	\centering
  \vspace{-5pt}
	\caption{
 Visual comparisons of D4IR and other methods for image dehazing on SOTS. Zoom in for a better view.
 }
	\label{fig_dehaze}
 \vspace{-10pt}
\end{figure*}

\subsection{Comparisons with the State-of-the-art \label{4.3}}
To validate the effectiveness of our D4IR, we provide quantitative and qualitative comparisons for image deraining, dehazing, and desnowing. 
Here, we mainly compare our D4IR with four kinds of methods: 
1) directly train the student network with the original training data of the teacher network (Student). 
2) distill the student network with the original degraded data without the GT supervision (Data). 
3) distill the student network by DFSR~\cite{zhang2021data} and DFMC~\cite{wang2024data}. 
Other data-free distillation methods are designed for high-level vision tasks, which cannot be applied to IR for comparison. 
4) the mainstream unsupervised methods that are trained on unpaired data. 

\noindent\textbf{For Image Deraining.} 
As shown in Tab. \ref{tab_derain}, it is observed that the performance of the student network obtained by our D4IR for image deraining improves by 0.91dB on PSNR and 0.023 on SSIM compared to ``Data".
This benefits from the wider range of data synthesized by our D4IR, which is domain-related to the original degraded data so as to facilitate the student network to focus on the knowledge of the teacher network more comprehensively.
Besides, the performance of our D4IR also far exceeds that of the GAN-based DFSR and performs better than DFMC (0.44dB and 0.024 higher on PSNR and SSIM). 
Moreover, D4IR also performs better than most mainstream unsupervised image deraining methods and achieves comparable performance with Mask-DerainGAN with only the half parameters. 
The visual comparisons in Fig. \ref{fig_rain} show that D4IR achieves a significant rain removal effect and is better than DFMC, DFSR, and students distilled with original data for removing rain marks. 

\noindent\textbf{For Image Dehazing.} 
As shown in Tab. \ref{tab_dehaze}, our D4IR also outperforms the student distilled with the original degraded data (0.04dB higher on PSNR and 0.001 higher on SSIM) and performs much better than DFSR and DFMC, which lack specific degradation-related losses. 
Besides, compared to the popular unsupervised image dehazing methods, D4IR has a much smaller number of parameters in second place on PSNR and SSIM. 
The visual result is given in Fig. \ref{fig_dehaze}. It shows that our D4IR has a significant dehazing effect and is closer to the GT than DFMC, DFSR, and ``Data".

\begin{figure}[t]
        \begin{minipage}{0.19\linewidth}
		\vspace{3pt}
		\centerline{\includegraphics[width=\textwidth]{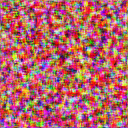}}
        \end{minipage}
        \begin{minipage}{0.19\linewidth}
		\vspace{3pt}
		\centerline{\includegraphics[width=\textwidth]{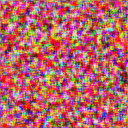}}
        \end{minipage}
        \begin{minipage}{0.19\linewidth}
		\vspace{3pt}
		\centerline{\includegraphics[width=\textwidth]{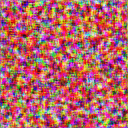}}
        \end{minipage}
        \begin{minipage}{0.19\linewidth}
		\vspace{3pt}
		\centerline{\includegraphics[width=\textwidth]{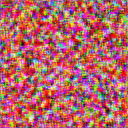}}
        \end{minipage}
        \begin{minipage}{0.19\linewidth}
		\vspace{3pt}
		\centerline{\includegraphics[width=\textwidth]{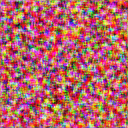}}
        \end{minipage}

        \begin{minipage}{0.19\linewidth}
		\centerline{\includegraphics[width=\textwidth]{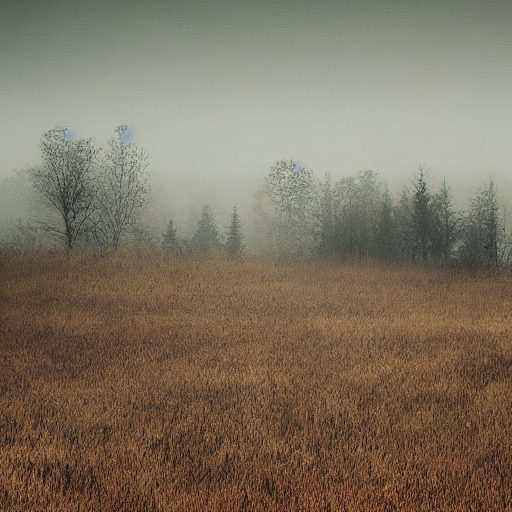}}
        \end{minipage}
        \begin{minipage}{0.19\linewidth}
		\centerline{\includegraphics[width=\textwidth]{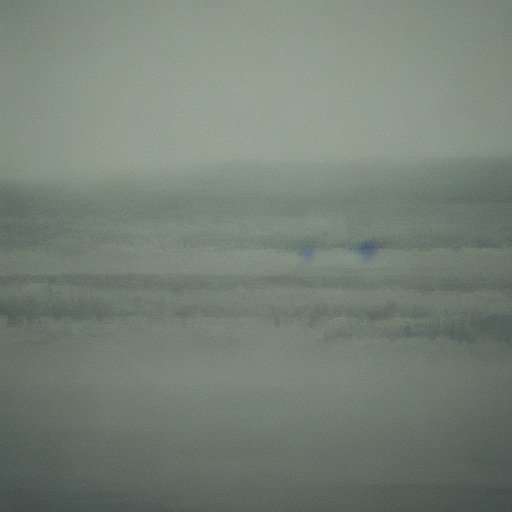}}
        \end{minipage}
        \begin{minipage}{0.19\linewidth}
		\centerline{\includegraphics[width=\textwidth]{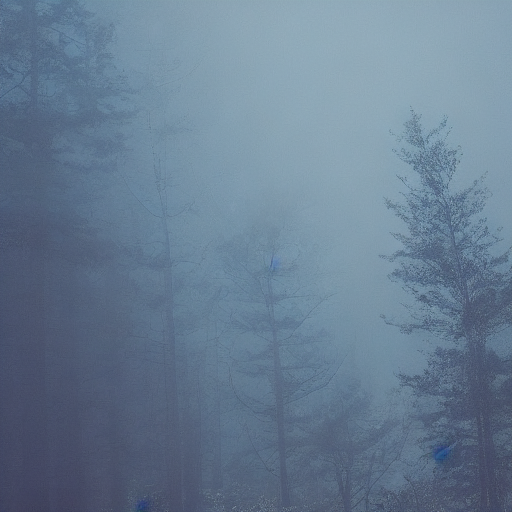}}
        \end{minipage}
        \begin{minipage}{0.19\linewidth}
		\centerline{\includegraphics[width=\textwidth]{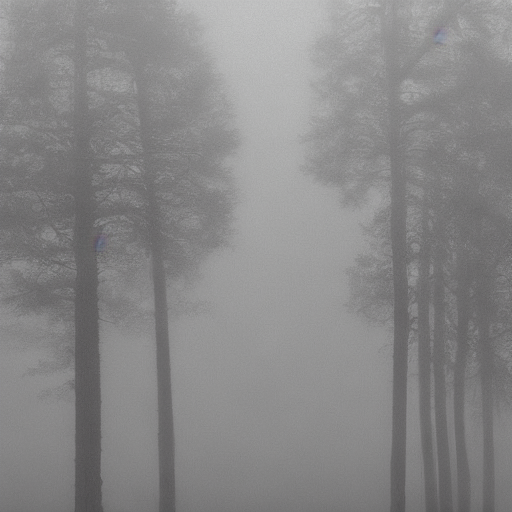}}
        \end{minipage}
        \begin{minipage}{0.19\linewidth}
		\centerline{\includegraphics[width=\textwidth]{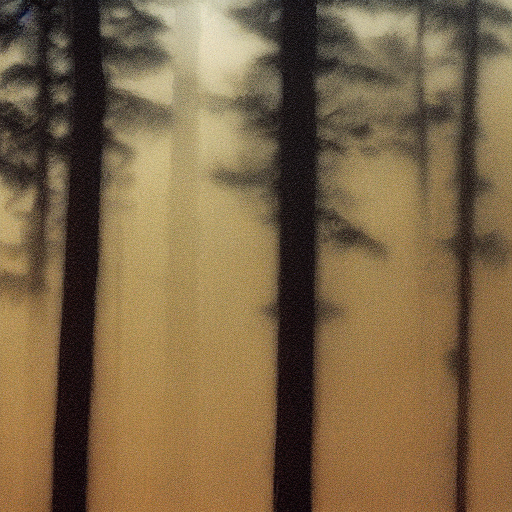}}
        \end{minipage}

        \begin{minipage}{0.19\linewidth}
		\vspace{3pt}
		\centerline{\includegraphics[width=\textwidth]{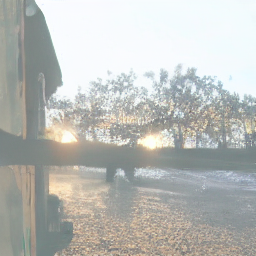}}
        \end{minipage}
        \begin{minipage}{0.19\linewidth}
		\vspace{3pt}
		\centerline{\includegraphics[width=\textwidth]{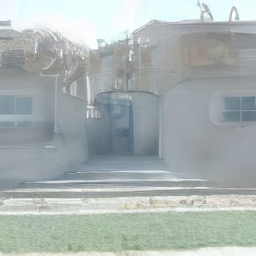}}
        \end{minipage}
        \begin{minipage}{0.19\linewidth}
		\vspace{3pt}
		\centerline{\includegraphics[width=\textwidth]{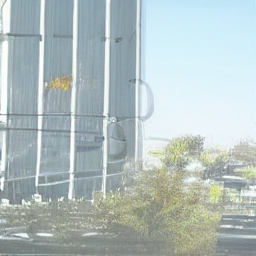}}
        \end{minipage}
        \begin{minipage}{0.19\linewidth}
		\vspace{3pt}
		\centerline{\includegraphics[width=\textwidth]{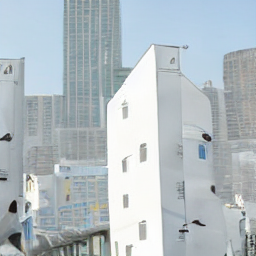}}
        \end{minipage}
        \begin{minipage}{0.19\linewidth}
		\vspace{3pt}
		\centerline{\includegraphics[width=\textwidth]{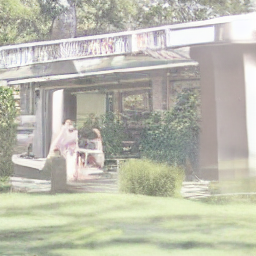}}
        \end{minipage}

	\centering
	\caption{Visualized samples synthesized by DFMC (Top), SD (Middle), and our D4IR (Bottom) for image dehazing. 
 }
    \vspace{-10pt}
	\label{fig_result}
\end{figure}


In Fig. \ref{fig_result}, we present visualized samples synthesized by DFMC, the pre-trained SD model, and our D4IR for image dehazing. The results indicate that GAN-based DFMC, which initiates from pure noise, struggles to produce images with semantic information. Additionally, generating images with rich texture and color details using simple textual prompts proves challenging for SD. In contrast, our D4IR method generates images with more detailed texture and semantic information compared to both DFMC and SD. 

The results for image desnowing are in the supplement.

\subsection{Ablation Studies \label{4.4}}
Here, we mainly conduct the ablation experiments on the image deraining task as follows: 

\noindent\textbf{Break-down Ablation.} We analyze the effect of the degradation-aware prompt adapter (DPA) and content-driven conditional diffusion (CCD) by setting different input $z_0$ (noise and CCD) and prompts (none, textual features same as SD, content features encoded from clean images, and DPA) for frozen SD model in Tab.~\ref{tab_ab}.
It is observed that the performance of M1 is slightly better than that of M2 since the ``text-to-image" generative model is powerful in generating images with original textual prompts. 
Besides, the degradation-aware prompts can not work well without content-related information (M2) for the absence of content information compared with M3. 
Both textual degradation prompts (M5) and our proposed DPA (D4IR) effectively improve student models' performance compared with none prompts (M4). 
Our D4IR performs the best by jointly utilizing DPA and CCD to generate images close to the original degraded data. It improves 1.65dB on PSNR compared with the model relying solely on the pre-trained SD model (M1) and 1.34dB on PSNR compared with the model directly distilled with the web-collected data (M0).

\begin{table}[t]
    \centering
    \small
    \begin{tabular}{ccccc}
    \toprule
         \textbf{Models} & \textbf{$z_0$} & \textbf{Prompt} & \textbf{PSNR(dB) $\uparrow$} & \textbf{SSIM $\uparrow$}  \\
    \hline
         M0 & $\times$ & $\times$ & 28.69 & 0.876 \\
         M1 & noise & text & 28.38 & 0.879 \\
         M2 & noise & DPA & 28.20 & 0.862 \\
         M3 & noise & content & 29.08 & 0.893 \\
         M4 & CCD & none & 29.02 & 0.888 \\
         M5 & CCD & text & 29.60 & 0.903 \\
         D4IR & CCD & DPA & 30.03 & 0.906  \\
    \bottomrule    
    \end{tabular}
    \caption{
    Break-down Ablation of D4IR on Rain100L.}
    \label{tab_ab}
\end{table}

\noindent\textbf{Real-world Dataset.} 
For further general evaluation in practical use, we conducted experiments on the real-world rainy dataset SPA \cite{wang2019spatial}. 
As shown in Tab. \ref{tab_derain_SPA2}, our D4IR  also has comparable performance with the student distilled with original data in real-world scenarios (0.08dB higher on PSNR). 
More comparisons with other unsupervised methods are presented in the supplementary material. 

\begin{table}[t]
    \centering
    \small
    \begin{tabular}{ccccc}
    \toprule
         \textbf{Method} & \textbf{Teacher} & \textbf{Student} & \textbf{Data} & \textbf{D4IR}  \\
    \midrule
         PSNR(dB) $\uparrow$ & 33.59 & 33.55 & 33.45 & 33.53 \\
         SSIM $\uparrow$ & 0.935 & 0.933 & 0.932 & 0.932 \\
    \bottomrule
    \end{tabular}
    \caption{D4IR for image deraining on SPA.}
    \label{tab_derain_SPA2}
    \vspace{-10pt}
\end{table}

\noindent\textbf{Different Backbones of Teacher Network.} 
We also validate D4IR with a transformer-based teacher backbone Restormer \cite{zamir2022restormer} on Rain100L. Due to resource constraints, we use Restormer with halved feature channels (from 48 to 24) as our teacher network and a quarter of feature channels (from 48 to 12) as the student network. 
As shown in Tab. \ref{tab_restormer}, the shrunk model capacity also leads to a large performance loss of the student network compared to the teacher network. 
Besides, it is observed that the performance of our D4IR is slightly lower than that of the student network distilled with the original degraded data. 
The reason lies in that the images generated by the diffusion model still differ from the real training data while the self-attention mechanism of the transformer pays more attention to the global contextual information of the images. 

\begin{table}[htbp]
    \centering
    \small
    \begin{tabular}{ccccc}
    \toprule
         \textbf{Method} & \textbf{Teacher} & \textbf{Student} & \textbf{Data} & \textbf{D4IR}  \\
    \midrule
         PSNR(dB) $\uparrow$ & 35.75 & 28.37 & 26.21 & 26.01 \\
         SSIM $\uparrow$ & 0.964 & 0.895 & 0.851 & 0.817 \\
    \bottomrule
    \end{tabular}
    \caption{D4IR based on Restormer for image deraining.}
    \label{tab_restormer}
    \vspace{-5pt}
\end{table}

\section{Conclusion}
This paper proposes a simple yet effective data-free distillation method with degradation-aware diffusion for MWIR. 
To achieve this, we mainly consider three concerns, including: 
1) investigate the application of the conditional diffusion model to solve the unstable training of the traditional GANs in data-free learning; 
2) introduce a contrast-based prompt adapter to extract degradation-aware prompts from collected degraded images; 
and 3) start diffusion generation from content-related features of collected unpaired clean images. 
Extensive experiments show that our D4IR obtains reliable student networks without original data by effectively handling the distribution shifts of degradation and content.
In future work, we will continue to study more effective prompt generation to enable efficient model learning.


\bibliography{d4ir}

\end{document}